\documentclass[conference]{IEEEtran}
\IEEEoverridecommandlockouts
\usepackage{cite}
\usepackage{amsmath,amssymb,amsfonts}
\usepackage{algorithm}
\usepackage{algpseudocode}
\usepackage{multirow}
\usepackage{url}
\usepackage{graphics}

\usepackage{graphicx}

\usepackage{textcomp}
\usepackage{xcolor}
\def\BibTeX{{\rm B\kern-.05em{\sc i\kern-.025em b}\kern-.08em
    T\kern-.1667em\lower.7ex\hbox{E}\kern-.125emX}}
\begin{document}

\title{Personalized Colorectal Cancer Survivability Prediction with Machine Learning Methods*\\
\thanks{*National Science Foundation REU program research}
}

\author{\IEEEauthorblockN{1\textsuperscript{st} Samuel Li}
\IEEEauthorblockA{\textit{Princeton University} \\
\textit{}\\
Princeton, NJ \\
seli@princeton.edu}
\and
\IEEEauthorblockN{2\textsuperscript{nd} Talayeh Razzaghi}
\IEEEauthorblockA{\textit{New Mexico State University} \\
\textit{}\\
Las Cruces, NM \\
talayehr@nmsu.edu}
}

\maketitle

\begin{abstract}
In this work, we investigate the importance of ethnicity in colorectal cancer survivability prediction using machine learning techniques and the SEER cancer incidence database. We compare model performances for 2-year survivability prediction and feature importance rankings between Hispanic, White, and mixed patient populations. Our models consistently perform better on single-ethnicity populations and provide different feature importance rankings when trained in different populations. Additionally, we show our models achieve higher Area Under Curve (AUC) score than the best reported in the literature. We also apply imbalanced classification techniques to improve classification performance when the number of patients who have survived from colorectal cancer is much larger than who have not. These results provide evidence in favor for increased consideration of patient ethnicity in cancer survivability prediction, and for more personalized medicine in general.
\end{abstract}


\section{Introduction}
Colorectal cancer, defined as cancer starting in the colon or the rectum, is among the most common types of cancer for both men and women. The probability of developing colon or rectum cancer is about 4.5\% for men and 4.15\% for women. In 2018, the National Cancer Institute estimates there will be 140,250 new diagnoses of colorectal cancer and over 50,630 deaths caused by these cancers in the United States \cite{colorectal-stats}. This makes colorectal cancer the third leading cause of cancer-related deaths in the country. A patient’s survival time is largely dependent on the state of their cancer at the time of diagnosis: the 5-year survival rate for people with stage I colon cancer is approximately 92\%, while the 5-year survival rate of stage IV colon cancers is approximately 11\% \cite{colorectal-survival-stats}.

Upon receiving their diagnosis, a patient will likely want to know their x-year survivability, defined as the probability that they will survive beyond x years. An estimate of their survivability based only on the stage of cancer does not account for all the personal factors specific to that patient, and will be far less accurate than a prediction model that takes into account factors like patient age, race, primary site, etc. While it is difficult for a doctor to consider many factors, machine learning algorithms can efficiently find patterns in large datasets of patient data. Given the vast amount of diagnostic data that is available and continuously being generated, there are many opportunities to use machine learning techniques to provide more personalized and accurate survivability predictions.

Several works have been studied the application of machine learning in cancer prognosis and survivability prediction. Delen et al.~used data mining methods to predict breast cancer survivability in 2004 \cite{delen2005predicting}. Since then, with the increased adoption of machine learning methods, researchers have applied a wide variety of algorithms to survivability prediction of many types of cancer. Studies include the use of deep learning  \cite{delen2005predicting, al2017survivability}, ensemble methods \cite{al2013colon}, and imbalanced classification techniques \cite{liu2009decision}. These studies predict survivability based on demographic and diagnostic information for each patient.

One approach to provide patients with more personalized healthcare is to take the patient's race into account. It is shown that people of different racial groups have varying levels of susceptibility and responses to different diseases. Causes for discrepancies may include anything from genetic differences to environmental influences to cultural factors. Currently, there is a significant disparity in colorectal incidence and mortality rates between different racial groups. From 2011 to 2015, the National Cancer Institute reported that the colorectal cancer incidence rate was 38.8 per 100,000 for the White population, while the Hispanic and black populations had incidence rates of 47.6 and 33.5 per the same-size population, respectively. In the same years, the White, Hispanic, and black populations had recorded mortality rates of 14.1, 11.5, and 19.4 per 100,000, respectively \cite{race-mortality}.

Our research aims to develop models which effectively predict colorectal cancer survivability and to use these models to compare feature importance between populations of different race. In this paper, we predict 2-year survivability for Hispanic and White colorectal cancer patients, and compare important features in determining survivability. Our choice of these two ethnic groups is justified by the observation that they were the most frequently appearing ethnic groups in our dataset. While ``Hispanic'' describes an ethnicity, and therefore the two groups are not automatically disjoint by official race and ethnicity records, most self-described Hispanic people consider their Hispanic background as part of their racial background \cite{hispanic_origin}. In this study, we consider a patient White if the related ``race encoding'' variable is White and the related ``Hispanic origin'' encoded their labels as Non-Hispanic.

Additionally, we investigate how we can improve upon the problem as an imbalanced classification problem. An imbalanced classification problem is defined as a problem in which the instance labels, in our case whether or not a patient survives beyond two years, are disproportionate. 
We apply imbalanced classification techniques to improve classification performance.

We will create a separate dataset for each race and train classifiers to predict survivability on each separated dataset. The trained models provide feature rankings among the patient features (age at diagnosis, tumor size, primary site, etc.), which allow feature comparison between Hispanic and White patients.

This paper is organized as follows: In Section \ref{dataSource}, we briefly introduce our dataset for this study. Section \ref{intro} discusses our methodology and its components as well as our data source. Results are reported in Section \ref{results}, and we discuss our findings in Section \ref{discussion}. We conclude our study in Section \ref{conclusion}.

\section{Data Source} \label{dataSource}
For this study, we use the diagnostic data available in the Surveillance, Epidemiology, and End Results (SEER) database \cite{seer-dataset}. Sponsored by the National Cancer Institute, SEER is the principal repository for cancer incidence and survival data in the United States. The SEER database contains over 10 million diagnostic records from 1973-2015. These records are collected from cancer registries across the country, and include the data from over ten types of cancer. For this study, we aggregated patient data from all geographical registries, and filtered our data to only include White and Hispanic patients. The mixed dataset is  formed by combining the Hispanic and White datasets. Table~\ref{dataset-info} shows the survivability distribution by ethnicity.

\begin{table}[h]
	\centering
	\caption{Description of label distribution by ethnicity of SEER data}
	\begin{tabular}{cccc}
	\hline
		\textbf{Ethnicity} & \textbf{Not Survived} & \textbf{Survived} & \textbf{Total} \\
		\hline
		White              & 53343 (19.4\%)   & 222204 (80.6\%)        &  275547  \\
		Hispanic           & 6762 (18\%)       & 30813 (82\%)           &  37575 \\
		\hline
	\end{tabular}
	
	\label{dataset-info}
\end{table}

section{Methodology}
\label{intro}
Our machine learning pipeline consists of three stages: 1) data preprocessing, 2) model training, and 3) model evaluation. Data preprocessing is performed to convert the raw data into a format that can be used as input to our models. Next, we train an array of data mining models using the preprocessed data. Finally, we evaluate each model through a variety of performance measures. Further details of these stages are presented in the next subsections. 

\subsection{Data Preprocessing} \label{dataPreprocess}
In this study, we use SEER data from the years 2004-2015 due to the following reasons. First, there is a major variable recode in the data beginning in 2004, which may have compromised data consistency if we had used patient data both prior and after 2004. Second, using relatively recent data is ideal because medical technology is continuously developing and improving, resulting in more timely, accurate diagnoses and better treatments. We use colorectal cancer patient data from all available registries.

Raw SEER data is encoded in pure ASCII files. These files are converted via Python scripts to a tabular format, from which we create two distinct datasets: one containing only Hispanic patients, and the other containing only White patients. We perform this separation at this point,  so that further modifications remain specific to each race.

We refer to instances and predictions, where a patient survives beyond two years as ``positive,'' and instances and predictions, where a patient dies of colorectal cancer before two years as ``negative.'' Accordingly, an instance is labeled negative if the survival time in months field is less than 24, and the cause of death is colon or rectum cancer. All instances not satisfying those two conditions are labeled positive.

Out of 133 features provided by SEER, we select the 20 features detailed in Table~\ref{variable-info} based on feature importance reported in the literature  \cite{delen2005predicting, al2013colon}. We select 16 categorical features and 4 continuous features. Each feature is carefully examined for outliers, missing values, and other inconsistencies. For the age diagnosed, number of malignant tumors, and months survived features, we delete the entire instance if the feature is missing. For the tumor size and number of positive nodes features,  where there are a large proportion of missing or unknown values, we employ the Multivariate imputation by chained equations (MICE) imputation technique described shortly. It is worth noting that many other related studies cited in this paper have dropped large proportions of instances with missing data. However, we aim at minimizing instance eliminations because the SEER dataset does not have abundantly many Hispanic patient instances to begin with.

\begin{table}[h]
	\centering
	\caption{Predictor features for survivability prediction models}
	\tiny
	\begin{tabular}{ll}
  	    \hline
		\textbf{Categorical features} & \textbf{Unique values}  \\
		\hline
		Marital status    & 7                       \\
		Sex               & 2                       \\
		Primary site      & 13                      \\
		Histology         & 139                     \\
		Behavior          & 2                       \\
		Grade             & 5                       \\
		Diagnostic confirmation  & 8                       \\
		Extension         & 65                      \\
		Lymph Nodes       & 18                      \\
		Metastasis        & 25                      \\
		Tumor size evaluation  & 7                       \\
		Node evaluation        & 7                       \\
		Metastasis evaluation  & 7                       \\
		Surgery site      & 34                      \\
		Reason no surgery & 8                       \\
		Summary stage     & 5                       \\
		Registry          & 18                      \\
		\hline
	\end{tabular}
	\quad
	\centering
	\begin{tabular}{lll}
	    \hline
		\textbf{Continuous features} & \textbf{Mean} & \textbf{S.D.}  \\
		\hline
		Age diagnosed     & 68.3          & 14             \\
		Positive nodes    & 1.57          & 4.26           \\
		Number of tumors  & 1.4           & 0.717          \\
		Tumor size        & 43.1          & 37.3          \\
		\hline
	\end{tabular}
	
	\label{variable-info}
\end{table}

In order to reconcile categorical and continuous features, we convert each categorical feature into multiple binary variables with the one-hot encoding scheme. In one-hot encoding, one binary variable is created for each category of the categorical variable. The binary variable corresponding to the original category value of the instance has a value of 1, and the other binary variables have 0 has their values. For example, for each instance the marital status categorical feature would be converted to seven new binary variables. In an instance where the patient was married, the binary variable corresponding to the married category would have a value of 1, and the other six binary variables would have values of 0. It is noteworthy that one-hot encoding can result in very sparse input vectors when the number of possible categories in some variables is relatively large, as is the case for our data.

Multivariate imputation by chained equations (MICE) is among the most common strategies for handling missing values in electronic health data \cite{azur2011multiple}. MICE initially fills all missing values with the mean or median of the corresponding non-missing features, and then iteratively cycles through each feature requiring imputations and performs regression by treating all other features as independent variables. Algorithm \ref{mice-algo} provides pseudocode for the MICE algorithm.
	
\begin{algorithm}
	\hspace*{\algorithmicindent} \textbf{Data:} input data; \\ \hspace*{\algorithmicindent} \textbf{numcycles:} number of cycles to run MICE (usually 10)
	\begin{algorithmic}[1]
		\Function{MICE}{Data, numcycles}
			\State Perform mean/median imputation on all missing values in dataset
			\For{i in numcycles}
				\For {feature $f_m$ in set of features with any missing values}
					\State Regression(Data - $column_{f_m}$, $column_{f_m}$)
					\State Replace originally missing features in $column_{f_m}$ with regression result
				\EndFor
			\EndFor
		\EndFunction
		
	\end{algorithmic}
	\caption{MICE Imputation}
	\label{mice-algo}
\end{algorithm}

\subsection{Model Training} \label{Models}
For this study, we employ the following classifiers to predict survivability. Our choice of these classifiers is justified based on our preliminary test performance as well as models reported in literature.
\begin{itemize}
	\item \textit{Logistic Regression} is a classifier that evaluates the weighted sum of the input components, and applies the ``sigmoid'' function to the weighted sum. The output from the sigmoid function can be interpreted as the probability of the positive class in the binary classification scenario. The weights are learned during training.
	\item \textit{Neural Network} is composed of layers of interconnected nodes. Each node receives a weighted sum of the outputs from all the nodes in the previous layer. This value is then adjusted by a biad value assiengned to the node. Next, an activation function is applied to the adjusted value, and the result is the output of the node. The weights in each layer are determined by the backpropagation training method, which updates each weight based on its influence on the loss function. Neural networks are at the forefront of recent deep learning progress \cite{lecun2015deep}.
	 \item \textit{Random Forest} is an {\it ensemble} of decision trees, where each tree splits nodes based on random features (instead of best features) \cite{breiman2001random}. The final output class is the class which receives the majority vote over the individual trees. To reduce the effects of overfitting, the method splits the features randomly.
	\item \textit{AdaBoost} \cite{freund1999short} is an {\it ensemble} of sequentially trained classifiers. Each instance in the training set is initialized with equal weight. Mis-classified instances are given higher weight while training the next classifier. The final classifier is a weighted model on all the classifiers, where each is weighted by its accuracy during training. In this study, the decision stump is used as the base classifier.
	\item \textit {Imbalanced Classification} techniques are used to deal with imbalanced data. We have employed the following techniques to improve model performances.
    \begin{itemize}
    \item \textit{Random Undersampling}
    uses a fraction of the majority class instances randomly, to make the proportion between class instances more balanced \cite{he2008learning}.
    \item \textit{Cost-sensitive learning}
    penalizes the model severely for mislabeling any instance of the minority class.
   
	\end{itemize}

\end{itemize}

\subsection{Model Evaluation} \label{modelEval}
As our class labels are imbalanced, the conventional metric of accuracy is an inadequate indicator of model performance: to see this, observe that any model, which always predicts that the patient survives, would achieve over 80\% accuracy. The most common evaluation metric in the literature is Area Under Curve (AUC) of the ROC curve. The AUC score is intuitively interpreted as the probability that, given a positive and a negative instance, the positive instance is ranked more likely to be positive than the negative instance. The AUC score can be calculated given a set of predicted labels and the corresponding true labels. We employ AUC as the primary evaluation metric, so that our results can be more comparable to results reported in the literature.

In addition to AUC, we also calculate the G-mean as an indicator for our performance on an imbalanced classification problem. To obtain the G-mean, we first define the following four metrics.
\begin{itemize}
\item \text{TP: the number of positive instances labeled positive}
\item \text{TN: the number of negative instances labeled negative}
\item \text{FP: the number of negative instances labeled positive}
\item \text{FN: the number of positive instances labeled negative}
\end{itemize}
The confusion matrix shown in Table~\ref{confusion-matrix} summarizes the definitions.

\begin{table}[h]
	\caption{Confusion matrix for binary classification}
	\centering
	\begin{tabular}{llll}
		&                                   & \multicolumn{2}{c}{True Class}                     \\
		\cline{3-4}
		&                                   & Not Survived            & Survived                 \\
		\cline{3-4}
		\multirow{2}{*}{{Predicted}} & \multicolumn{1}{|l|}{Not Survived} & \multicolumn{1}{l|}{TN} & \multicolumn{1}{l|}{FN}  \\
		\cline{3-4}
		\multicolumn{1}{l}{}                & \multicolumn{1}{|l|}{Survived}     & \multicolumn{1}{l|}{FP} & \multicolumn{1}{l|}{TP}  \\
		\cline{3-4}
		&                                   &                         &
	\end{tabular}	
	\label{confusion-matrix}
\end{table}

\noindent The G-mean is defined as follows:
\begin{align*}
	\text{G-mean} = \sqrt{\text{sensitivity}\times \text{specificity}},
\end{align*}
where sensitivity and specificity are defined as
\begin{align*}
\text{sensitivity} = \frac{TP}{TP + FN}, \quad\text{specificity} = \frac{TN}{TN + FP}.
\end{align*}
The G-mean is a viable metric for imbalanced classification because it weights the rate of correctly predicting the majority and minority classes equally.

\section{Results}
\label{results}
We trained our models with Scikit-learn, an open-source Python machine learning library. We used Scikit-learn version 0.19.1 and Python 3.6.4. All experiments were executed on a Linux machine with an Intel i7-3700 3.4 GHz processor and 16GB RAM running Ubuntu 18.04 LTS.

We used 5-fold cross-validation to determine the best hyperparameters for each model, and recorded results on the test sets, each composed of a randomly-selected 20 percent of the data. Table~\ref{auc-scores} shows the AUC results for each model and dataset. Our AUC scores are calculated on each model's prediction of the probability that the true label is positive. This will provide a higher AUC score because wrong predictions are not penalized as heavily if their probability prediction was closer to 0.5. The AUC scores we achieved with our proposed Neural Network are above the best colorectal cancer prediction scores reported in literature of 0.8675 by \cite{al2017survivability}, which was also produced by neural networks for 2-year colorectal cancer survivability. However, we emphasize that direct score comparison is not fair since our data was pre-screened by ethnicity; rather, we demonstrated that we can achieve comparable performance on much smaller datasets by providing more personalized predictions. Our proposed Neural Network architecture consists of: 3 fully-connected hidden layers of 400 neurons, ReLU activation for hidden layers, 0.1 dropout between each layer besides input layer, and sigmoid activation for the output layer.
\begin{table}[h]
	\centering
	\caption{AUC results for trained models on the test set for three populations: Hispanic, White, and combination of both Hispanic and White (Mixed)}
	\begin{tabular}{llll}\hline
		\textbf{Model}      & \textbf{Hispanic} & \textbf{White} & \textbf{Mixed}  \\
		\hline
		Logistic Regression & 0.859              & 0.872           & 0.870             \\
		Random Forest       & 0.855              & 0.865           & 0.849            \\
		AdaBoost            & 0.859              & 0.871           & 0.859            \\
		Neural Network       & 0.873              & 0.875           & 0.856 \\
        \hline
	\end{tabular}
	
	\label{auc-scores}
\end{table}

Table~\ref{factor-rankings} ranks the top seven features for each model and dataset. Feature ranking models are obtained from the coefficient and parameter values of the trained models, in the context of each model. In the logistic regression model, a larger magnitude coefficient for a feature indicates that the feature has relatively greater impact on the label prediction. For decision tree-based models like random forest and AdaBoost, we use Gini impurity and entropy scores for each feature, which reveal the feature that provide the most information when a decision tree node is split on that feature. Since our models are trained on one-hot encoded features, we add the feature importance values for all the one-hot encoded columns corresponding to a single original feature (histology, surgery site, etc.) to obtain the score used to list feature importance (from high importance to less importance).

\begin{table}[h]
	\centering
	\caption{Feature rankings for each model and ethnic population}
	\resizebox{\columnwidth}{!}{%
	\begin{tabular}{ll|ll|ll}\hline
		\multicolumn{2}{c}{\textbf{Logistic Regression}}         & \multicolumn{2}{c}{\textbf{Random Forest}}              & \multicolumn{2}{c}{\textbf{AdaBoost}}                   \\ 		\hline
		\multicolumn{1}{l}{\textbf{Hispanic}} & \textbf{White}   & \multicolumn{1}{l}{\textbf{Hispanic}} & \textbf{White}  & \multicolumn{1}{l}{\textbf{Hispanic}} & 			\textbf{White}  \\ \hline
		Histology                             & Histology        & Metastasis                            & Metastasis      & Extension                             & Extension       \\
		Extension                             & Lymph node inv.  & Stage                                 & Stage           & Histology                             & Age             \\
		Metastasis                            & Extension        & Age                                   & Age             & Age                                   & Histology        \\
		Surgery site                          & Surgery site     & No surg. reason                       & No surg. reason & Tumor size                            & Positive nodes  \\
		Diagnostic conf.                      & Metastasis       & Positive nodes                        & Positive nodes   & Positive nodes                        & Metastasis      \\
		Lymph node inv.                       & Diagnostic conf. & Surgery site                          & Surgery site    & Metastasis                            & Tumor size      \\
		No surg. reason                       & Primary site     & Tumor size                            & Tumor size      & Surgery site                          & Surgery site \\\hline
	\end{tabular}%
	}
	
	\label{factor-rankings}
	
\end{table}

\noindent Table~\ref{gmean-compare} shows the results calculated from the cost-sensitive learning and undersampling methods. Random undersampling balances the two classes through decreasing the size
of the majority class, and in cost-sensitive learning, the weight of each instances in minority class is five times as the majority class.
Random undersampling method is applied to the datasets prior to model training. Our results are comparable to the 0.792 G-mean reported by Al-Bahrani and Agrawal \cite{al2017survivability}. We observe that the undersampled logistic regression yields slightly better performance through building separate models for each ethnicity.

\begin{table}[h]
	\caption{G-mean results for selected models and ethnic populations. Bold numbers denote the best
G-mean results compared to Al-Bahrani and Agrawal \cite{al2017survivability}.}
	\centering
	\begin{tabular}{llll} \hline
		\textbf{Model}               & \textbf{Hispanic} & \textbf{White} &   \\ \hline
		Logistic Regression          & 0.628              & 0.683           &   \\
		Cost-Sensitive Logistic Regression & 0.783              & \bf{0.800}             &   \\
		Undersampled Logistic Regression & 0.790 & \bf{0.800} & \\
		Random Forest                & 0.623              & 0.631           &   \\
		Cost-Sensitive Random Forest       & 0.782              & 0.775           &  \\
		Undersampled Random Forest & 0.787 & \bf{0.796} \\ \hline
	\end{tabular}
	\label{gmean-compare}
\end{table}

\section{Discussion}
\label{discussion}
Our results provide insight into the importance of patient ethnicity in cancer survivability prediction, as well as the cancer survivability prediction problem itself. From the AUC score Table~\ref{auc-scores}, we observed that models trained on mixed-ethnicity patient data performed consistently worse than those trained on single-ethnicity data. Logistic regression was the only model in which the mixed dataset yielded better results either the Hispanic or White datasets. The White dataset had significantly better results than the mixed dataset for all models, while the Hispanic dataset had slightly better results. This is very likely due to the Hispanic dataset being far smaller than the White dataset, rather than the Hispanic dataset being intrinsically harder to train on.

Our feature rankings results show fairly consistent variable rankings between the Hispanic and White datasets for the same model, with minor differences. For example, lymph node involvement is the second most important factor for White patient instances in the logistic regression model, but the sixth most important factor for Hispanic patient instances. It is expected that there are not dramatic differences in feature rankings between the two groups. However, even minor differences can be of use in the analysis of a patient's diagnostic data; for example, by giving an observation of a feature more consideration for a patient with a certain race. Additionally, we observe in Table~\ref{factor-rankings} that different models have widely varying feature rankings. This confirms that many features are not independent, and that there are many viable ways to make survivability predictions with competitive results. We note that neural networks are unable to provide feature rankings due to the non-transparent nature of the model.

From the imbalanced classification results, we observe that the logistic regression and random forest models both perform better with respect to the G-mean metric with imbalanced classification techniques. Both weighted classes and undersampling improved the G-mean significantly. Applying these techniques to the AdaBoost and neural network model achieves similar results (Table~\ref{more-scores}). The same table shows that G-mean scores greatly improved, while AUC scores slightly increased after applying undersampling to AdaBoost, and slightly decreased after applying undersampling to a neural network. Applying imbalanced classification methods demonstrates the trade-off between sensitivity and specificity, and the inherent hardness of this classification problem. In the original models sensitivity and specificity values were around 0.9 and 0.4, respectively. After applying either weighted classes or undersampling, sensitivity and specificity values were both close to 0.8.

\begin{table}[h]
	\centering
\caption{G-mean and AUC results of the AdaBoost and Neural Network for imbalanced classification}
	\begin{tabular}{lllll}
		& \multicolumn{2}{c}{\textbf{Hispanic }} & \multicolumn{2}{c}{\textbf{White}}  \\ \cline{2-3} \cline{4-5}
		\textbf{Model}              & \textbf{G-mean} & \textbf{AUC}         & \textbf{G-mean} & \textbf{AUC}      \\ \hline
		AdaBoost                    & 0.627           & 0.859                & 0.671           & 0.859             \\
		Undersampled AdaBoost       & 0.787           & 0.865                & 0.798           & 0.871             \\
		Neural Network              & 0.600             & 0.873                & 0.700             & 0.875             \\
		Undersampled Neural Network & 0.788           & 0.868                & 0.796           & 0.873             \\
\hline
	\end{tabular}
	\label{more-scores}
\end{table}

The trade-off implies that instances with similar feature values have different labels. In some instances this is due to both actual survival times being close in value but on opposite sides of the 24-month cutoff. In other instances, survival times can actually differ widely, suggesting that we do not have all the relevant information. Indeed, this is the case: we only have the colorectal cancer diagnostic information and basic personal information. We hypothesize that more extensive data which includes features like personal health history and lifestyle information would improve performance beyond the best currently reported in literature. However, there is currently no large-scale public dataset with such degree of patient information.

\section{Conclusion}
\label{conclusion}
In this study we applied machine learning methods to predict 2-year colorectal cancer survivability. Specifically, we trained logistic regression, random forest, AdaBoost, and neural network models on SEER data we carefully preprocessed. Our models achieved AUC score of 0.87, improving upon the best-reported score in the literature. We used our models to investigate the impact of ethnicity on model performance, as well as diagnostic feature rankings. Our models achieved superior performance on single-ethnicity patient data compared to the mixed-ethnicity data, and we also observed minor differences in feature importance. Finally, we used the imbalanced classification methods of cost-sensitive learning and undersampling to improve the G-mean metric to approximately 0.8.

We infer from our results that cancer survivability prediction is a difficult classification problem, and that the SEER dataset may not contain enough information to significantly improve the performance of our current models. Future work includes extending our focus on ethnicity beyond the Hispanic and White patient groups, as well as to other types of cancer.

\section{Supplementary Material}
The authors provided an implementation of the proposed method at \url{https://github.com/samuelli97/cancer_survivability}.

\section*{Acknowledgment}
This work was funded in part by National Science Foundation's Research Experiences for Undergraduates Program (ACI~\#1559723).

\end{document}